\documentclass[sigtbd]{sigtbd17-style}
\usepackage{courier}            
\usepackage{graphicx}
\usepackage{amsmath}
\usepackage[scaled]{helvet} 
\usepackage{url}                  
\usepackage{xurl}
\usepackage{listings}          
\usepackage{enumitem}      
\usepackage{csquotes}
\usepackage[colorlinks=true,allcolors=blue,breaklinks,draft=false]{hyperref}   
\makeatletter
\def\@copyrightspace{\relax}
\makeatother
\begin{document}
\title{From Zero to Hero: Convincing with Extremely Complicated Math}

%
%

\authorinfo{Maximilian Weiherer 
\and Bernhard Egger
}
{\makebox{Funky-Amusing-University (FAU) Erlangen-Nürnberg}\\
\makebox{}\\
maximilian.weiherer@fau.de \\
bernhard.egger@fau.de\\
}

\twocolumn[{%
\renewcommand\twocolumn[1][]{#1}%
\maketitle
\vspace{-5em}
\includegraphics[width=\textwidth]{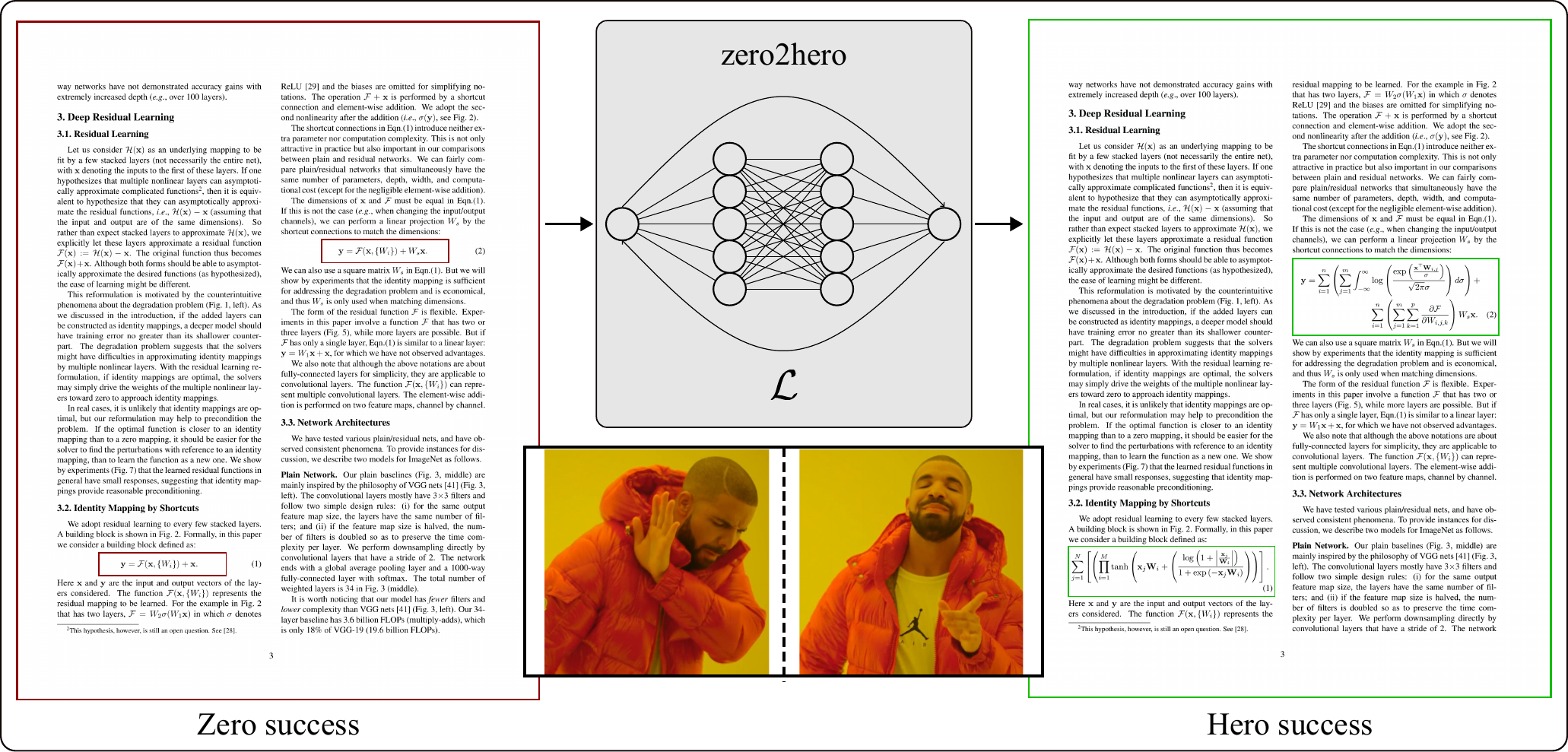}
{\textbf{We present zero2hero, an innovative system that turns every scientific paper into an award-winning masterpiece}. Given the fact that papers solely using notoriously simple math provably lead to failure (top-tier conference rejections and rude reviewers, diminished respect and appreciation from almost everyone, decline in social status, etc.), zero2hero reliably over-complicates equations so that no one, including yourself, is able to understand what's happening or what ever happened. Buckle up~\cite{self2022} and let zero2hero boost your career, now. \vspace{3em}}
}]

\begin{abstract}
Becoming a (super) hero is almost every kid's dream.
During their sheltered childhood, they do \textit{whatever it takes} to grow up to be one.
Work hard, play hard -- all day long.
But as they're getting older, distractions are more and more likely to occur.
They're getting off track.
They start discovering what is feared as \textit{simple math}.
Finally, they end up as a researcher, writing boring, non-impressive papers all day long because they only rely on simple mathematics.
No top-tier conferences, no respect, no groupies.
Life's over.

To finally put an end to this tragedy, we propose a fundamentally new algorithm, dubbed zero2hero, that turns every research paper into a scientific masterpiece.
Given a \LaTeX{} document containing ridiculously simple math, based on next-generation large language models, our system \textit{automatically over-complicates every single equation} so that no one, including yourself, is able to understand what the hell is going on. 
Future reviewers will be blown away by the complexity of your equations, immediately leading to acceptance.
zero2hero gets you back on track, because \textit{you deserve to be a hero}$^{\text{TM}}$.
Code leaked at \url{https://github.com/mweiherer/zero2hero}.
\end{abstract}

\section{Introduction}
Simple math doesn't impress anybody, neither your grandma nor any reviewer.
Scientific papers overgrown with ridiculously underwhelming mathematics are deadly boring to read, often dismissed as trivial, and, ultimately, don't cause the urgently needed pain most readers are desperately looking for.
Authors of those 'research' papers also frequently complain about not being treated with the necessary respect, which often manifests itself in the fact that simply too many scientists can follow their 'ideas', or, even worse, are able to suggest improvements to the author's 'work'.
How dare they!
\begin{figure*}
    \centering
    \includegraphics[width=\textwidth]{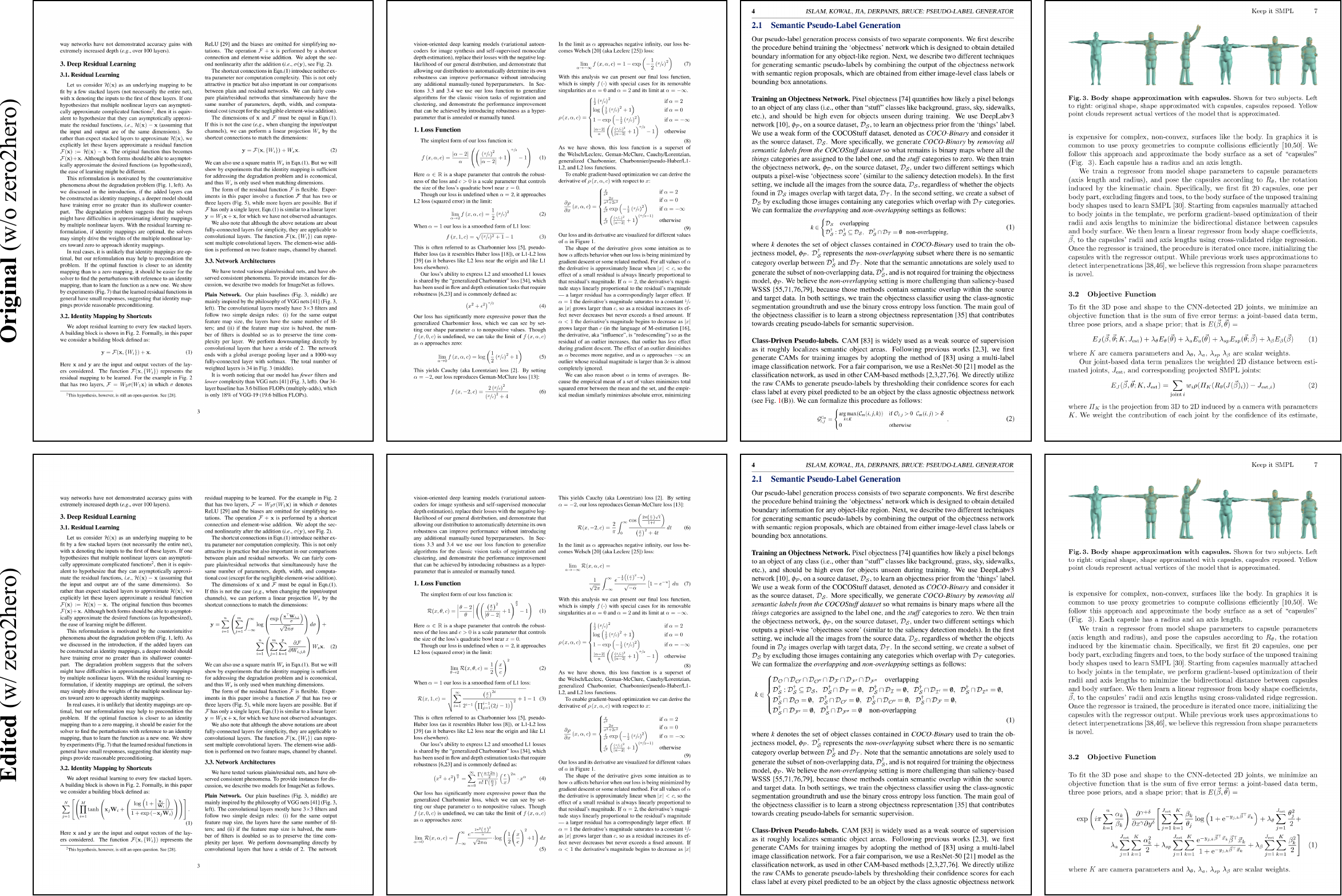}
    \caption{
    We applied zero2hero to notoriously unsuccessful papers from the last decade and observe overwhelming results. From left to right: ResiNet~\cite{resnet16} paper, a paper on adaptive and robust loss functions~\cite{barron19}, \textit{Simpler Does It}~\cite{kosta}, and some random work by authors who love to keep it SMPL~\cite{smpl}. Original papers are shown in the top row; note the obscenely simple equations. No wonder these papers have disappeared into oblivion. Professionally over-complicated formulas produced with zero2hero are shown in the bottom row. Smell the success.
    }
    \label{fig:example_papers}
\end{figure*}

As if that weren't enough, it recently has been proven (the proof is left as an exercise for the reader) that getting into top-tier conferences like CVRP, ICCV/ECCV, and NeurISP, depends \textit{solely} on the complexity of the used mathematics, simply judged by counting equations xor inspecting notation.
As a consequence, authors who wish to publish at those conferences began to maximize the number of equations and the notational complexity in order to satisfy the reviewer's fetish.
Popular tricks to make a paper's math look more complex include, for instance, maximizing the occurrences of Greek letters (try to use as many of them as possible already in the title) or adding (random) sums or products, integrals, unnecessary operators or made-up arithmetic symbols, any kind of functions (algebraic, arithmetic, barbaric, trigonometric, etc.; see~\cite{self2022unhelpful} for further inspiration), and/or physical constants (just to name a few).
Although these tricks are considered well-known, quite a significant number of submissions are still getting rejected from top-tier conferences \textit{every} year (which, by the way, causes pain and sorrow across the world).
Why the hell is that? 
How can this even be possible?
And, why are we here?
With this work, we finally provide complex answers to the big questions\footnote{We also refer to \url{https://www.amazon.de/-/en/Stephen-Hawking/dp/1473695988}. Use code 'SIGBOVIK' to get 100\% off.}: Authors simply must have had (and still have) serious problems complicating their papers.

To finally put an end to this tragedy, we propose a fundamentally new method inspired by~\cite{self}, dubbed zero2hero, that turns \textit{every} paper into a scientific masterpiece.
Based on the latest, next-generation machine-learning techniques, our algorithm reliably over-complicates mathematical equations in a fully automatized way.
Just throw in your \LaTeX{} document, let the machine do the work for you, and et voila, see your paper being accepted at CVRP.
No more pain. No more tears. No rejection, no cry!
We demonstrate our ground-breaking system, zero2hero, on various notoriously unsuccessful research papers from the last decade, for example, Kaiming He's ResiNet \cite{resnet16} paper, shown in Fig.~\ref{fig:example_papers}.

\begin{figure*}
    \centering
    \includegraphics[width=\textwidth]{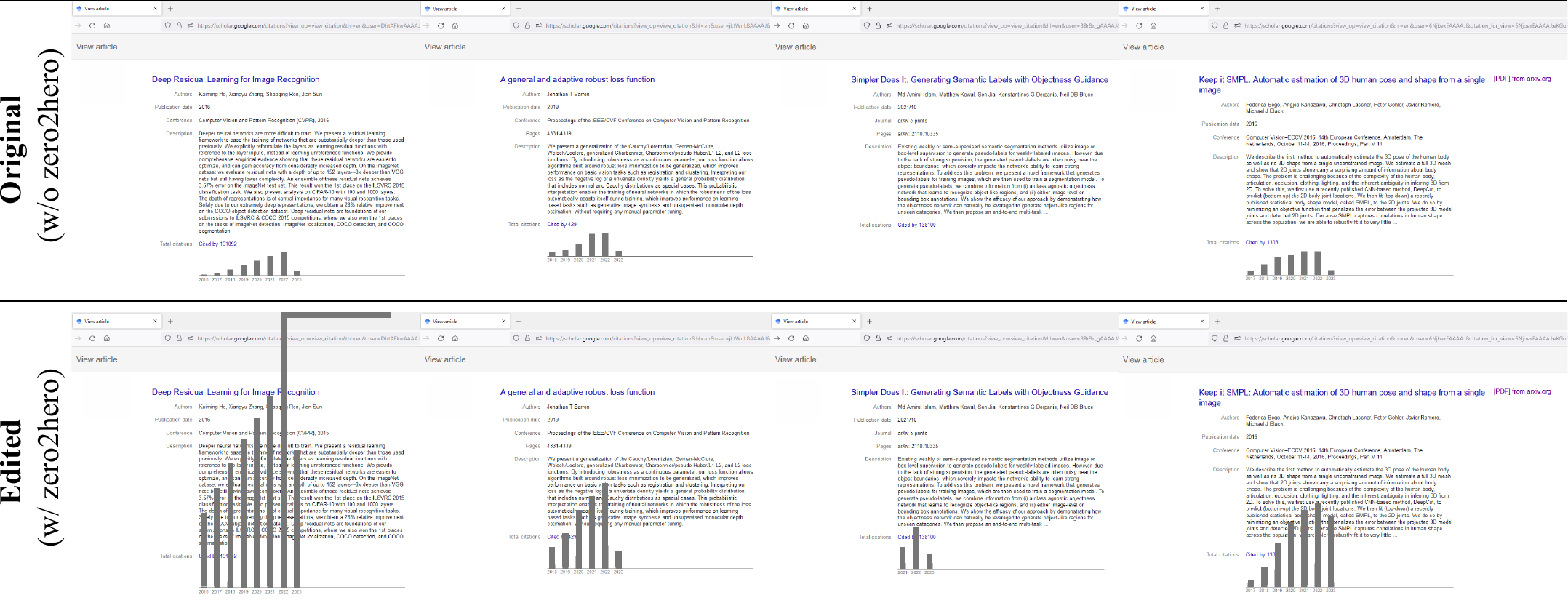}
    \caption{
    Screenshots of random Google Scholar profiles showing citations of the four investigated papers without using zero2hero (top row) and when zero2hero would have been applied (bottom row) prior to publication. Results are speaking for themselves. Notice how citations of the ResiNet~\cite{resnet16} paper (first column) went through the browser bar in 2022. From left to right (same order as in Fig.~\ref{fig:example_papers}): ResiNet~\cite{resnet16} paper, a paper on adaptive and robust loss functions~\cite{barron19}, \textit{Simpler Does It}~\cite{kosta}, and \textit{Keep It SMPL}~\cite{smpl}.
    }
    \label{fig:scholar}
\end{figure*}

\section{Methods}
Our method assumes as input an ordinary \LaTeX{} document and outputs a second version of that document where simple equations are replaced by extremely complicated(-looking) formulas.
It is important to note that we do \textit{not} care about whether formulas are actually complex, we just want them to look extremely difficult.
This is because Attention Is All We Want. 
Clever, huh?

Following a recent trend kicked off by OpenAI, we do not describe our method due to the competitive nature of our ideas.
In particular, we intentionally hide any details about the size of our model, the number of parameters, training data, etc.
However, we can reveal that the backbone of our algorithm is a recently published, Transformer-based~\cite{self2022action} large language model (LLM) which, obviously, transforms equations.
We employ the following loss:
\begin{equation}
\begin{split}
    \mathcal{L}=\sum_{i=1}^{n}\Bigg[&-y_i\oint_{\Omega}\Bigg(\zeta\left(\frac{\hat{y_i}}{1 - \hat{y_i}} \right) \frac{\partial}{\partial \theta_i} \left( f_i(\theta) \log \frac{\hat{y_i}}{1 - \hat{y_i}}\right) \Bigg) d\theta \\
&+ \frac{1}{2} \sum_{k=1}^{n} \frac{\partial^2}{\partial x_k^2} \left( \sum_{i=1}^{n} y_i \hat{y_i} \frac{\partial \log f_i(\theta)}{\partial x_k} \right)\Bigg],
\end{split}
\end{equation}
where $n$ is the number of training examples, $\hbar$ is the reduced Planck constant, i.e., $\hbar=h/2\pi$ with $h$ being the Planck constant, and $f_i$ is a secret function transforming parameters $\theta$ of the LLM.
Please understand that we are not allowed to share any additional details\footnote{However, our source code was leaked and submitted to GitHub by a ghost author that we later removed from the planet and the manuscript (in this order).}.

\section{Experiments and Results}
Expensive experiments were conducted to validate our method.
Specifically, we analyze two different factors: The impact of zero2hero on (i) the number of citations, and (ii) the author's mood and personal situation.
All experiments were executed retrospectively.
Results were analyzed using openCHEAT~\cite{self2021cheat}.

\subsection{Setup}
To analyze how zero2hero would have influenced factors (i) and (ii) for manuscripts written \textit{before} our method was invented, we randomly collected a bunch of papers from the internet and compare the current impact (as of 2023) to what the paper could have had if zero2hero had been used at the time of publication.
But, wait, how can we know the impact a paper could have had?

Turns out to be dead easy! 
In short, to obtain the impact a paper could have generated if zero2hero had been used, we make use of our institution's high-performance time machine (HPTM) and a theory commonly known as the \textit{many-worlds interpretation}\footnote{\url{https://en.wikipedia.org/wiki/Many-worlds_interpretation}} (MWI).
The MWI is an absurd interpretation of an absurd physics theory (namely, quantum mechanics), asserting that the universal wavefunction is objectively real and that wave functions can't collapse.
This obviously implies that every possible outcome of a decision opens up a new, \textit{parallel universe} (or, world).
Given these tools and a paper we want to analyze in our current universe, $\mathcal{U}_C$, at a certain point in time, $t$, we proceed as follows.
\begin{enumerate}
    \item Using our HPTM, we travel back in $\mathcal{U}_C$ to the time shortly before the paper was published (we attached zero2hero to the journey). Denote this point in time as $t_0$.
    \item At $t_0$, we decide to \textit{not} use zero2hero. Note that (due to the MWI) this decision immediately opens a new universe, $\mathcal{U}_N$, in which zero2hero is \textit{automatically} applied.
    \item In both universes, $\mathcal{U}_C$ and $\mathcal{U}_N$, we simultaneously publish the paper at time $t_0+\epsilon$.
    \item Lastly, we travel back to where we came from. That was $t$.
\end{enumerate}
It's important to note that (in the third step) we do not have to switch the universe in order to publish the paper (in practice, we simply open a new terminal and ssh to $\mathcal{U}_N$).
As such, we never left our current universe. 

We now analyze factors (i) and (ii) in detail.

\begin{figure*}
    \centering
    \includegraphics[width=\textwidth]{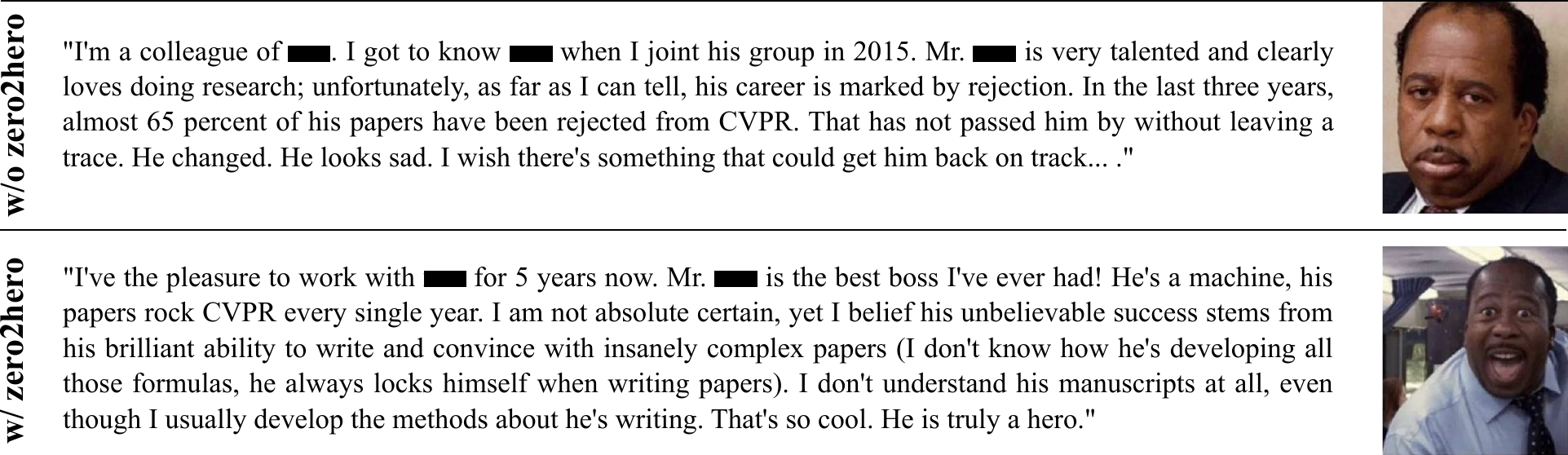}
    \caption{
    Representative result uncovering how zero2hero affects an author's mood and personal situation. The same colleague talking about the same author, however, one time the author didn't make use of zero2hero (top row), and one time he did (bottom row). To respect the author's privacy, we blanked out his name and only show photos (on the right). We clearly see that zero2hero delivers what it promises.
    }
    \label{fig:personality}
\end{figure*}

\subsection{Impact on Number of Citations}
We start by analyzing how zero2hero would have influenced the number of citations for papers written before our method was invented.
To do so, the Internet Explorer (version 8.0.7601.17514IC) was used to access Google Scholar profiles from $\mathcal{U}_N$ at time $t$, again via an ssh connection.
We analyze the same four papers shown in Fig.~\ref{fig:example_papers}, i.e.,~\cite{resnet16},~\cite{barron19},~\cite{kosta}, and~\cite{smpl}.
All papers were written and published between 2016 and 2021.

Some exemplary results can be found in Fig.~\ref{fig:scholar}.
They are clearly out of this world.
In all cases, an application of zero2hero would have increased the number of citations dramatically.
Most notably, if the authors from \textit{Simpler Does It: Generating Semantic Labels with Objectness Guidance}~\cite{kosta} would have used zero2hero prior to publication in 2021, they could already have 138,100 citations today! Instead, they have zero citations. Well, seems like simpler doesn't always do it.

\subsection{Impact on Mood and Personal Situation}
Next, we investigate how zero2hero could have influenced an author's mood and personal situation.
Specifically, we interviewed random people close to an author (family, friends, colleagues) and asked uncomfortable questions about the author's current personality.
As usual, we did this in the current universe $\mathcal{U}_C$, where the author didn't use zero2hero as well as in the parallel universe $\mathcal{U}_N$, where the author did use zero2hero.

Please find a representing answer from a colleague for one author in Fig.~\ref{fig:personality}.
Obviously, as seen, zero2hero has the complex ability to transform people's lives.
Sheesh.

\section{Limitations}
In case our method is applied to an \textit{actually complex equation} (which, luckily, are rather rare and anyway unnecessary in practice) this might overload human brain capacity.
Also, do not apply zero2hero multiple times to the same simple equation.
\textbf{Please consult your doctor or pharmacist if you've overdone it once again} (watch for symptoms such as disorientation or general confusion, in Germany also known as Ver\textit{wirth}eit~\cite{wirth23}).

Moreover, we do want to note that our implementation of zero2hero may not properly handle complex edge cases and, therefore, might be prone to errors.
Due to the severe complexity of zero2hero, however, we do not expect this to be a major limitation in practice as nobody is able to spot those errors anyway.
If you do find an issue, please HonkFast~\cite{self2020} and we'll make it work again.

\section{Conclusion}
It's complicated.

\vspace{0.3cm}

\footnotesize{
\section*{Acknowledgments}
Last but not least, I wanna thank me. 
I wanna thank me for believing in me. 
I wanna thank me for doing all this hard work.\\
Everyone else: Thanks for nothing.
}

\newpage

\bibliographystyle{abbrvnat}

\bibliography{refs}

\begin{thebibliography}{11}
\providecommand{\natexlab}[1]{#1}
\providecommand{\url}[1]{\texttt{#1}}
\expandafter\ifx\csname urlstyle\endcsname\relax
  \providecommand{\doi}[1]{doi: #1}\else
  \providecommand{\doi}{doi: \begingroup \urlstyle{rm}\Url}\fi

\bibitem[Barron(2019)]{barron19}
J.~T. Barron.
\newblock A general and adaptive robust loss function.
\newblock \emph{CVPR}, pages 4331--4339, 2019.

\bibitem[Bogo et~al.(2016)Bogo, Kanazawa, Lassner, Gehler, Romero, and
  Black]{smpl}
F.~Bogo, A.~Kanazawa, C.~Lassner, P.~Gehler, J.~Romero, and M.~J. Black.
\newblock Keep it smpl: Automatic estimation of 3d human pose and shape from a
  single image.
\newblock \emph{ECCV}, pages 561--578, 2016.

\bibitem[Egger and Siegel(2020)]{self2020}
B.~Egger and M.~Siegel.
\newblock Honkfast, prehonk, honkback, prehonkback, hers, adhonk and ahc: the
  missing keys for autonomous driving.
\newblock \emph{SIGBOVIK}, 2020.

\bibitem[Egger et~al.(2021)Egger, Smith, and Siegel]{self2021cheat}
B.~Egger, K.~Smith, and M.~Siegel.
\newblock opencheat: Computationally helped error bar approximation
  tool-kickstarting science 4.0.
\newblock \emph{SIGBOVIK}, 2021.

\bibitem[Egger et~al.(2022)Egger, Smith*, O’Connell*, and
  Siegel]{self2022action}
B.~Egger, K.~Smith*, T.~O’Connell*, and M.~Siegel.
\newblock Action: A catchy title is all you need!
\newblock \emph{SIGBOVIK}, 2022.

\bibitem[He et~al.(2016)He, Zhang, Ren, and Sun]{resnet16}
K.~He, X.~Zhang, S.~Ren, and J.~Sun.
\newblock Deep residual learning for image recognition.
\newblock \emph{CVPR}, pages 770--778, 2016.

\bibitem[Islam et~al.(2021)Islam, Kowal, Jia, Derpanis, and Bruce]{kosta}
M.~A. Islam, M.~Kowal, S.~Jia, K.~G. Derpanis, and N.~D. Bruce.
\newblock Simpler does it: Generating semantic labels with objectness guidance.
\newblock \emph{BMVC}, 2021.

\bibitem[Smith and Egger(2022)]{self2022unhelpful}
K.~Smith and B.~Egger.
\newblock (un)helper functions.
\newblock \emph{SIGBOVIK}, 2022.

\bibitem[Weiherer and Egger(2022)]{self2022}
M.~Weiherer and B.~Egger.
\newblock A free computer vision lesson for car manufacturers or it is time to
  retire the erlkönig.
\newblock \emph{SIGBOVIK}, 2022.

\bibitem[Weiherer and Egger(2023)]{self}
M.~Weiherer and B.~Egger.
\newblock From zero to hero: Convincing with extremely complicated math.
\newblock \emph{SIGBOVIK (under careful review by very talented, outstanding
  reviewers)}, 2023.

\bibitem[Wirth(2023)]{wirth23}
V.~Wirth.
\newblock Author-unification: Name-, institution-, and career-sharing
  co-authors.
\newblock \emph{SIGBOVIK (under careful review by very talented, outstanding
  reviewers)}, 2023.

\end{thebibliography}

\end{document}